\documentclass[letterpaper, 10 pt, conference]{ieeeconf}  % Comment this line out if you need a4paper

\usepackage{times}
\usepackage{latexsym}
\usepackage{url}
\usepackage{multirow}
\usepackage{booktabs}
\usepackage{graphicx}
\usepackage[normalem]{ulem}
\usepackage{fancyvrb}
\usepackage{verbatim} % for comment blocks
\usepackage{color}

\IEEEoverridecommandlockouts                              % This command is only needed if
\title{\LARGE \bf
A Multi-layer LSTM-based Approach for Robot Command Interaction Modeling
}

\author{Martino Mensio$^{1}$, Emanuele Bastianelli$^{1}$, Ilaria Tiddi$^{2}$, Giuseppe Rizzo$^{3}$% <-this % stops a space
\thanks{$^{1}$Knowledge Media Institute,
        The Open University, UK
        {\tt\small \{name.surname\}@open.ac.uk}}%
\thanks{$^{2}$Department of Computer Science,
		VU University Amsterdam, NL
        {\tt\small ilaria.tiddi@vu.nl}}%
\thanks{$^{3}$Istituto Superiore Mario Boella, Italy
        {\tt\small giuseppe.rizzo@ismb.it}}%
}

\begin{document}

\maketitle
\thispagestyle{empty}
\pagestyle{empty}

%%%%%%%%%%%%%%%%%%%%%%%%%%%%%%%%%%%%%%%%%%%%%%%%%%%%%%%%%%%%%%%%%%%%%%%%%%%%%%%%
\begin{abstract}
As the first robotic platforms slowly approach our everyday life,
%and advancements in Natural Language Understanding have reached convincing results,
we can imagine a near future where service robots will be easily accessible by non-expert users through vocal interfaces. The capability of managing natural language would indeed speed up the process of integrating such platform in the ordinary life. Semantic parsing is a fundamental task of the Natural Language Understanding process, as it allows extracting the meaning of a user utterance to be used by a machine. In this paper, we present a preliminary study to semantically parse user vocal commands for a House Service robot, using a multi-layer Long-Short Term Memory neural network with attention mechanism. The system is trained on the Human Robot Interaction Corpus, and it is preliminarily compared with previous approaches.
\end{abstract}

\section{Introduction}
The area of Natural Language Understanding (NLU) has been gaining a growing consensus in recent years, also thanks to the trends dictated by new voice assistants, e.g. Amazon Echo. In parallel, the first cost-accessible commercial robots are growing in number, e.g. iRobot Roomba. % and Flymo 1200R.
As platforms evolve to expose complex services, it is legit to expect that they will be integrated with NLU capabilities. Natural language is, in fact, one of the most powerful and flexible communication tools, and vocal interface will become a mandatory feature, if we want service robots to be accessible by a wider range of users, especially non-expert ones.

Semantic parsing, the process of extracting interpretations from natural language, is a fundamental brick in NLU. In the last twenty years, a body of works have proposed solutions to this problem for virtual or real autonomous agents. Several approaches have been followed, from grammar-based ones~\cite{BosOka:07alr,Kruiff:07www}, to purely statistical ones~\cite{Tellex:11aim, ChenMooney:11nlinstruction, kim:2013ACL}, as well as hybrid approaches~\cite{Matuszek:12rcs, artzi-zettlemoyer:13weaklysuper, Thomason:15ijcai}. Among these, the work in~\cite{bastianelli:16ijcai} (henceforth BAS16) fosters the reliance on established linguistic theories to represent semantics of actions expressed in user's commands. A SVM-based statistical semantic parser is trained over the Human-Robot Interaction Corpus~\cite{bastianelli:14lrec} (HuRIC), which represents an attempt of bridging between the NLU for robots and more linguistically- and cognitively-sound theories of meaning representation, namely Frame Semantics~\cite{Fillmore:85frames} and the related FrameNet~\cite{Baker:98framenet} resource.

Moving from BAS16, and following the successful trend in applying deep neural network in Semantic Parsing~\cite{DBLP:conf/acl/JiaL16} and Semantic Role Labelling tasks~\cite{dossantos2015classifying}, where encoder-decoder recurrent architectures have proven particularly effective~\cite{Zhou15:acl, Yang:17emnlp}, in this paper we propose a preliminary study of the application of a multi-layer Long-Short Term Memory (LSTM) network to parse robotic commands from the HuRIC resource. Moreover, we also aim at showing that using a multi-layer LSTM is a viable solution even in poor training condition, as HuRIC contains only 527 examples.

\section{Approach}
%In this section we  provide a description of the addressed tasks and  describe the multi-layer LSTM-based approach that can be employed for it.
\label{sec:approach}
%\subsection{The tasks}
%\label{subsec:task}
%IROS% The overall objective investigated in this work is the one of semantically parse transcriptions of vocal command for a real House Service robot. The produced interpretations are given in terms of the set of \textit{semantic frames} described in the HuRIC data set, which is used as training reference for our system. They, in turn, follow the formalisation given in FrameNet.

%IROS% Semantic frames are conceptual structures describing real world situations, such as actions, or more in general events, e.g. the action of \textit{Taking}. A set of \textit{frame elements} is also associated with each frame, specifying the entities in a sentence participating in defining the situation described by the frame, e.g. the \textsc{Theme}, which represents the object that is taken during a \textit{Taking} action. Specific words, called Lexical Units, work as hook between the text and the theory, e.g. the verb \textit{take} can ``evoke'' the frame \textit{Taking}.
The objective of the system is to semantically parse transcriptions of vocal commands for a House Service robot contained in the HuRIC data set. The outputs are given in terms of \textit{semantic frames}, which are conceptualisations of actions or more general events. Each frame is evoked in the text by a \textit{lexical unit} and, besides its type, it also includes a set of \textit{frame elements}, which represent the entities having a specific role in the situation described by the frame.
%IROS% Parsing semantic frames involves three actions. First, all the frames in a sentence need to be identified. In BAS16, the process is referred to as Action Detection (AD), since the frames considered are all actions. Vocal commands, in fact, express actions the user wants the robot to perform. This process traditionally takes the name of Frame Identification or Frame Prediction. Frame elements are extracted through a process of Semantic Role Labeling \cite{Carreras:05CoNLL}, which is composed by two sub-processes: Argument Identification (AI) and Argument Classification (AC). In the first, spans of frame elements in a sentence are identified for a given frame. In the second, a label representing the frame element type is assigned to each span. As an example, the interpretation of the command:
%IROS% TODO this example takes a lot of space, check the figure that describes the network together with the example
%IROS% As an example the command:
%IROS% $$take\: the\: book\: on\: the\: table$$
%IROS% would correspond to:
%IROS% $$[\mathbf{take}]_{Taking}\: [the\: book\: on\: the\: table]_{\footnotesize \textsc{Theme}}.$$
%IROS% It is worth noting that there is no constraint on whether any of the three tasks for parsing semantic frames needs to be done independently or jointly with the others.
%IROS% The three aforementioned tasks, namely AD, AI, AC, are also known in the domain of Conversational Agents as \textit{intent classification}, \textit{slot identification} and \textit{filling} respectively.
The tasks involved aim at structuring the input sentence into actionable information, through: \textit{i)} identifying the frame representing the user's willingness as a label for the whole sentence (the Action Detection task, AD); \textit{ii)} selecting the relevant text spans of frame elements in the sentence (the Argument Identification task, AI); \textit{iii)} assigning a type to each frame element span (the Argument Classification task, AI). %IROS% In the following, we will use these terms interchangeably.
%\subsection{Multi-layer LSTM Neural Network with Self-attention}
%\label{ssec:3layersLSTM}
\begin{figure}[!b]
\centering
\includegraphics[width=0.85\linewidth]{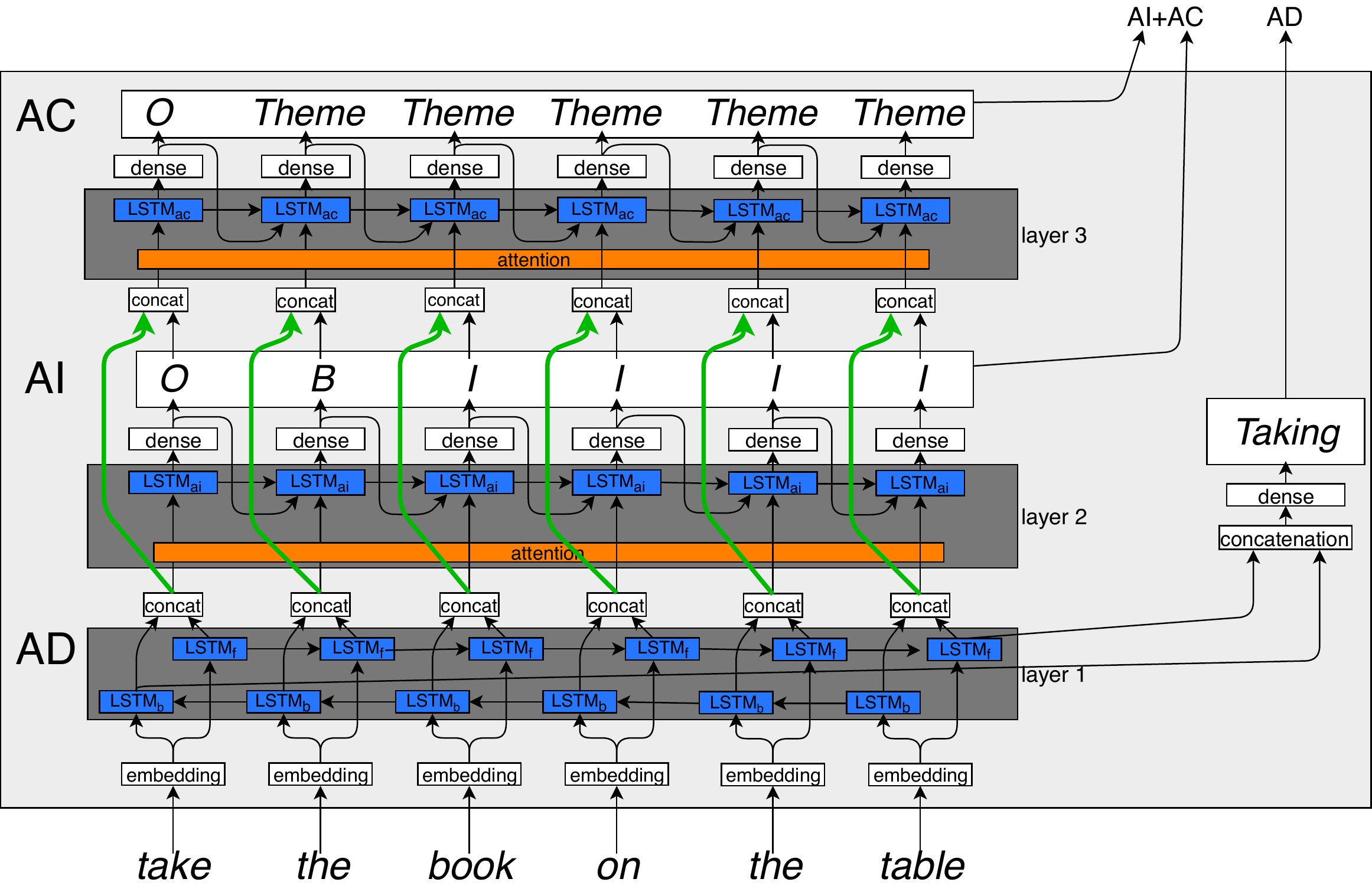}
\caption{The proposed 3-layer LSTM with self-attention network that addresses the three tasks of AD, AI and AC.}\label{fig:approach}
\end{figure}

Our approach builds upon the work in
%Inspired by the LSTM-based idea of
~\cite{Liu:16interspeech}, where a two-layer LSTM is applied to the Spoken Question Answering domain, which presents some similarities with our problem. We thus propose and test two variants of a multi-layered LSTM architecture, one with two layers (2L), exactly as in~\cite{Liu:16interspeech}, and one with three layers (3L). Both of them are fed with sequences of pre-trained word embeddings.
%take as input a sequence of words represented with pre-trained word embeddings.

In the 2L setting, the first layer is a bidirectional LSTM which is used to perform the AD task, while the second, a LSTM decoder with label dependencies~\cite{dupont2017label}, is used to perform jointly the AI and AC tasks. The bi-directionality of the first layer should capture backward dependencies, which are crucial for frame classification.
In the 3L configuration, instead, we introduce a third layer to divide the tasks of AI and AC. Every layer in the network thus solves one of the three semantic parsing tasks, as illustrated in Figure~\ref{fig:approach}.
%Inspired by the three different steps of semantic parsing described in \ref{subsec:task}, and , our idea is to introduce a third layer to divide the tasks of AI and AC.
The first two layers are still based on~\cite{Liu:16interspeech}, but the second one only predicts the \texttt{IOB} labels (\texttt{I}nside, \texttt{O}utside or \texttt{B}eginning of an element, \cite{ratnaparkhi:98phd}) for the AI rather than including also information about frame element types. The third layer takes as input the outputs of the AI layer, i.e. the \texttt{IOB} labels, combined with the internal representation from the first layer through \textit{highway connections}~\cite{srivastava2015highway} (green connections in Figure~\ref{fig:approach}), and outputs labels for the frame element types.
For each layer we make use of a self-attention mechanism~\cite{bahdanau2014neural} that learns weights, which enable combination of word-level features.
The loss function used to train the network corresponds to the sum of the cross-entropies ($H_x$) of each task, namely $ loss = H_{AD}+H_{AI}+H_{AC}$.

%IROS% These connections, underlined in green in the Figure, are known  as \textit{highway connections}~\cite{srivastava2015highway} and serve as shortcuts around the second layer. Highway connections are crucial in the network structure, as  they provide the third layer with semantic information required to perform the correct classification.
%The combination between the AI and AC labels occurs externally to the network, namely by taking the \texttt{IOB} label produced  during the AI step, and using a majority vote system over the AC labels, relying on word spans identified by the second layer. This operation can fix some errors in the third layer.
% majority vote given by the words inside a given span identified by the second layer
%IROS% To train the network the Adam Optimizer is used to perform stochastic gradient descent. The minimisation function is a clipped gradient that results as the sum of three loss functions, one for each pair of target and prediction type (AD, AI and AC), all of them summarised as the cross-entropy between target values and predicted values as follows:
%IROS% $$ loss = H_{AD} + H_{AI} + H_{AC} $$
%IROS% where H$_{x}$ represents the cross-entropy of a single processing step. In the specific, H$_{AD}$ is calculated as the softmax cross-entropy between labels and classification, while both H$_{AI}$ and H$_{AC}$ are evaluated as a weighted cross-entropy loss for a sequence of predictions.

\section{Experimental Evaluation}
\label{sec:exp}
% {\color{red}First, we perform a quantitative analysis of our system for semantic parsing, through comparing our results with the approach described in BAS16. ( First, we perform an analysis of our system in terms of semantic parsing. Secondly we evaluate the performances on the whole interpretation chain.)}

%IROS% This section reports the experimental evaluation carried out on our approach to parse vocal commands for a House Service robot, through comparing our results with the approach described in BAS16.
The Human-Robot Interaction Corpus (HuRIC) is used as data set for training, evaluation and comparison. It comprises 527 annotated sentences corresponding to vocal commands given to a robot in a house environment. FrameNet-style annotations are provided over each sentence, for a total of 16 frame types and an average of 33 examples for each frame. Hyper-parameter tuning, training and testing is performed through a 5-fold evaluation schema. Each network configuration has been tested  with (ATT) and without (NO-ATT) self-attention layers. Word embeddings have been pre-trained with GloVe~\cite{pennington2014glove} over the Common Crawl resource~\cite{common:11}.

\begin{table}[!htb]
  \small
  \centering
  \caption{F-measure of the three single stages of Semantic parsing (AD, AI, AC) and of the Whole Interpretation Chain.}
  %3L/2L stands for the number of layers used, while ATT/NO-ATT reflects the usage of attention}
    \begin{tabular}{r|ccc|c|}
          & {AD} & {AI} & {AC} & {Whole Chain}\\
\hline
    BAS16 &  94.67\% & 90.74\% & \textbf{94.93}\% & 41.70\% \\
\hline
    3L-ATT    & 94.44\% & 94.73\% & 94.69\% & 43.67\%\\
    3L-NO-ATT & 95.37\% & \textbf{94.90}\% & 91.90\% & 41.92\%\\
    2L-ATT    & \textbf{96.29}\% & 94.40\% & 92.30\% & \textbf{44.54}\%\\
    2L-NO-ATT & 94.44\% & 94.50\% & 92.45\% & 42.79\%\
    %IROS% \hline
    %IROS% TODO remove these two last rows? Not essential?
    %IROS% %3L-ATT-W    & 94.44\% & 94.44\% & 94.44\% & 95.70\% & 89.45\% & 92.47\% & 94.62\% & 87.13\% & 90.72\% \\
    %IROS% %2L-ATT-W    & 96.29\% & 96.29\% & 96.29\% & 95.67\% & 88.94\% & 92.19\% & 94.59\% & 86.63\% & 90.43\% \\
    \end{tabular}%
  \label{tab:res}%
\end{table}%

%\subsection{Evaluation Protocol}
%...
\subsection{Semantic Parsing}
%IROS% In BAS16, the three steps of semantic parsing (i.e. Action Detection, Argument Identification and Argument Classification) are implemented as separated processes composing a pipeline. Each process relies on semantic information coming from the previous steps, e.g. the AI and AC make use of the information about the frames evoked in a sentence. The results reported in BAS16 assume only gold semantic information for each processing step, i.e. AI and AC take gold frames as input, and AC takes also gold argument boundary annotations. Unfortunately, we cannot compare directly with this scenario, as our network performs the three steps concurrently. Implicit semantic representations are encoded and decoded automatically, while the information flows forward and backward in the network. It is thus impossible to provide gold semantic information to an intermediate step: that would require, for instance, to know the values of a gold encoding of a sentence with the correct tagged frame. This issue affects only the AI and AC steps, as the AD does not depend on any previous semantic information.
The first evaluation is performed on the semantic parsing tasks. In BAS16, the three steps (AD, AI, AC) are implemented by three independent blocks chained in a pipeline, and therefore they can be evaluated independently by using gold information. Instead, in our approach the information between the layers are implicit and gold values cannot be emulated.
%\sout{To overcome this problem, we propose a first run of evaluation, called LSTM-GS, in which gold semantic information is provided as an explicit feature together with each word, simulating how such information is provided in the representation adopted by BAS16.} {\color{red} HERE WE NEED SOME CLARIFICATION ABOUT HOW WE SOLVE THIS PROBLEM FOR AI AND AC}.
For this reason, the measures reported in Table~\ref{tab:res} for the AI and AC are computed only on the portion of examples which are correctly classified by the preceding step.
%IROS% , e.g. AI is evaluated only on examples which are correctly classified by the AD step, while AC evaluated only on examples correctly classified by both AD and AI.
In this way, it is possible to estimate the performance of the three different tasks independently, assuming gold information coming from the previous steps. %IROS% Precision and Recall scores have been therefore evaluated according to the following criterion for the AI and AC tasks:

From Table~\ref{tab:res}, we can see that the LSTM performs well in the AD task, with best results for the 2L-ATT setting. Every LSTM configuration outperforms BAS16 in the AI task. On the contrary, only the 3L-ATT configuration behaves similarly to BAS16 in the AC. The attention mechanism appears here to be crucial, as the scores drop significantly without it. This comes from the fact that the attention enables the third layer to better focus on the whole span identified by the AI task, with a softer alignment model. The AC is more complicated than the other two tasks, and the scarcity of examples seems to be a discriminant factor in these settings.

\subsection{Whole Interpretation Chain}
%\noindent \textbf{Whole Interpretation Chain}
The second experimental setting aims at evaluating the whole interpretation chain, from the transcribed user utterance to the grounded robot command.
%IROS% In BAS16, three different runs are reported: a first one, where both correct transcriptions and gold morpho-syntactic information are considered; a second one, in which the morpho-syntax is provided by the Stanford CoreNLP over correct transcriptions; and a third one, where transcriptions are directly obtained from the audio commands using the Google Speech-To-Text Cloud API\footnote{https://cloud.google.com/speech-to-text/}. Here, we compare only with the second run, thus assuming correct transcription coming from an ASR module. As mentioned before, it is not possible in our case to experiment gold morpho-syntactic information, as this is implicitly encoded by the embedding layer. Moreover, HuRIC provides only audio files and correct transcriptions. Using the Google Speech-To-Text API to transcribe them today would certainly result in different outcomes with respect to BAS16 due to improvement of the models.
Performances are evaluated on the fully-grounded robot commands. Each frame extracted from an utterance needs to be not only linguistically instantiated, but its arguments have also to be grounded in the environment.
%IROS%: nouns referring to objects or areas have to be linked to the correct entities in the environment.
Each vocal command in HuRIC is paired with a semantic map representing the environment where the command has been given. A command counts as correctly grounded when all the frames in the related sentence are correctly instantiated, and all the frame entities are linked to the proper entities in the semantic map. Please refer to Section 4 of BAS16 for an in-depth definition of this task.

In Table~\ref{tab:res}, under the Whole Chain column, we compare with the ``Gold transcr., CoreNLP'' run of BAS16, because our preliminary comparison wants to test the system over correct speech recognition transcriptions. Notice that our approach presents structural differences with BAS16, e.g. we do not need any morpho-syntactic parsing (CoreNLP), as we use word embeddings to represent words. While all the network configurations perform better than BAS16, the one that reaches highest results is 2L-ATT, which has also the best score on the AD task. This follows from the fact that the correct grounding of a command primarily depends on the correct interpretation of the frames contained in it. Misclassifying a frame but correctly recognising the frame elements, on the other hand, compromises the whole result.
%the one that has the best score on the AD task, namely the 2L-ATT as correctly classifying a frame appears to more important both in terms of grounding.
%{\color{red}As we can see from Table~\ref{tab:res}, the configuration that reaches the best performances is the one that has the best score on the AD task, namely the 2L-ATT, as correctly classifying a frame appears to be much more important both in terms of grounding and in terms of human judgement.}

\section{Conclusions}
\label{sec:conclusion}

%IROS% Natural (Spoken) Language Understanding is an area which has been growing in the recent years, especially thanks to the rapid development of modern voice digital assistants. We expect a similar development towards vocal robotic interfaces to happen, as soon as robots will be more ready to accomplish complex tasks in more sophisticated scenarios. The capability to understand and respond to natural language commands will be a crucial feature in future service robots, especially in the perspective of non-expert users.

In this paper, we presented a preliminary study to semantically parse natural language robotic commands from the HuRIC resource using a multi-layer LSTM network with attention layers. For our initial tests, we compared with the work in~\cite{bastianelli:16ijcai}, showing that a LSTM-based approach is a viable solution also in such a poor training condition (only 527 examples in HuRIC, $\sim$33 examples per frame). Future works should cover the study of the attentions values to better explain the network behaviour. Such information could be used to adjust the interpretation process through some dialogue with the user. Finally, a mechanism to embed perceptual information in the LSTM framework should be investigated, as fostered in~\cite{bastianelli:16ijcai}.

\bibliographystyle{IEEEtran}
\bibliography{thebibliography}

\end{document}